\documentclass{llncs}
\usepackage{makeidx}  

\usepackage[utf8]{inputenc} 
\usepackage[T1]{fontenc}    
\usepackage{hyperref}       
\usepackage{url}            
\usepackage{booktabs}       
\usepackage{amsfonts}       
\usepackage{nicefrac}       
\usepackage{microtype}      
\usepackage[english]{babel}
\usepackage{amsmath}
\usepackage{dblfloatfix}
\usepackage{array}
\usepackage{caption}
\usepackage{float}
\usepackage{algorithmicx}
\usepackage{algorithm}
\usepackage{algpseudocode}
\usepackage{amssymb}
\usepackage{graphicx}
\usepackage{setspace}
\usepackage[colorinlistoftodos]{todonotes}
\usepackage[normalem]{ulem}

\usepackage[font={small}]{caption}
\usepackage{layouts}
\usepackage{soul}
\begin{document}
\pagestyle{headings}  

\title{Optimization for Large-Scale Machine Learning with Distributed Features and Observations}
\titlerunning{Doubly Distributed Optimization}  
%
\author{Alexandros Nathan \and Diego Klabjan}
%
%
%
\institute{  Department of Industrial Engineering and Management Sciences \\ 
 Northwestern University, Evanston IL, 60208, USA \\
\email{\href{mailto:anathan@u.northwestern.edu}{anathan@u.northwestern.edu}}, \email{\href{mailto:d-klabjan@northwestern.edu}{d-klabjan@northwestern.edu}}
}

\maketitle              

\begin{abstract}
As the size of modern data sets exceeds the disk and memory capacities
of a single computer, machine learning practitioners have resorted to parallel and distributed computing. Given that optimization is one of the pillars of machine learning and predictive modeling, distributed optimization methods have recently garnered ample attention in the literature. Although previous research has mostly focused on settings where either the observations, or features of the problem at hand are stored in distributed fashion, the situation where both are partitioned across the nodes of a computer cluster (doubly distributed) has barely been studied. In this work we propose two doubly distributed optimization algorithms. The first one falls under the umbrella of distributed dual coordinate ascent methods, while the second one belongs to the class of stochastic gradient/coordinate descent hybrid methods. 
We conduct numerical experiments in Spark using real-world and simulated data sets and study the scaling properties of our methods. Our empirical evaluation of the proposed algorithms demonstrates the out-performance of a block distributed ADMM method, which, to the best of our knowledge is the only other existing doubly distributed optimization algorithm.

\end{abstract}

\section{Introduction}

The collection and analysis of data is widespread nowadays across many industries. As the size of modern data sets exceeds the disk and memory capacities of a single computer, it is imperative to store them and analyze them distributively. Designing efficient and scalable distributed optimization algorithms is a challenging, yet increasingly important task. There exists a large body of literature studying algorithms where either the features or the observations associated with a machine learning task are stored in distributed fashion. Nevertheless, little attention has been given to settings where the data is doubly distributed, i.e., when both features and observations are distributed across the nodes of a computer cluster. This scenario may arise in practice as a result of distinct data collection efforts focusing on different features -- we are assuming that the result of each data collection process is stored using the split across observations. The benefit of using doubly distributed algorithms stems from the fact that one can bypass the costly step (due to network bandwidth) of moving data between servers to avoid the two levels of parallelism.



In this work, we propose two algorithms that are amenable to the doubly distributed setting, namely  D3CA (Doubly Distributed Dual Coordinate Ascent) and RADiSA (RAndom Distributed Stochastic Algorithm). These methods can solve a broad class of problems that can be posed as minimization of the sum of convex functions plus a convex regularization term (e.g. least squares, logistic regression, support vector machines). 

D3CA builds on previous distributed dual coordinate ascent methods \cite{jaggi2014communication,ma2015adding,yang2013trading}, allowing features to be distributed in addition to observations. The main idea behind distributed dual methods is to approximately solve many smaller sub-problems (also referred to herein as partitions) instead of solving a large one. Upon the completion of the local optimization procedure, the primal and dual variables are aggregated, and the process is repeated until convergence. Since each sub-problem contains only a subset of the original features, the same dual variables are present in multiple partitions of the data. This creates the need to aggregate the dual variables corresponding to the same observations. To ensure dual feasibility, we average them and retrieve the primal variables by leveraging the primal-dual relationship \eqref{eq:primal-dual}, which we discuss in section ~\ref{Algorithms}.


In contrast with D3CA, RADiSA is a primal method and is related to a recent line of work \cite{mokhtari2016doubly,wang2014randomized,zhao2014accelerated} on combining Coordinate Descent (CD) methods with Stochastic Gradient Descent (SGD). Its name has the following interpretation: the randomness is due to the fact that at every iteration, each sub-problem is assigned a random sub-block of local features; the stochastic component owes its name to the parameter update scheme, which follows closely that of the SGD algorithm. The work most pertinent to RADiSA is RAPSA \cite{mokhtari2016doubly}. The main distinction between the two methods is that RAPSA follows a distributed gradient (mini-batch SGD) framework, in that in each global iteration there is a single (full or partial) parameter update. Such methods suffer from high communication cost in distributed environments. RADiSA, which follows a local update scheme similar to D3CA, is a communication-efficient generalization of RAPSA, coupled with the stochastic variance reduction gradient (SVRG) technique \cite{johnson2013accelerating}. 

The contributions of our work are summarized as follows:

\begin{itemize}
    \item We address the problem of training a model when the data is distributed across observations and features. We propose two doubly distributed optimization methods.
    \item We perform a computational study to empirically evaluate the two methods. Both methods outperform on all instances the block splitting variant of ADMM \cite{parikh2014block}, which, to the best of our knowledge, is the only other existing doubly distributed optimization algorithm.
\end{itemize}

The remainder of the paper is organized as follows: Section ~\ref{Related_Work} discusses related works in distributed optimization; Section ~\ref{Algorithms} provides an overview of the problem under consideration, and presents the proposed algorithms; in Section ~\ref{NumExperiments} we present the results for our numerical experiments, where we compare D3CA and two versions of RADiSA against ADMM. 

\section{Related Work}

\label{Related_Work}

\paragraph*{Stochastic Gradient Descent Methods} SGD is one of the most widely-used optimization methods in machine learning. Its low per-iteration cost and small memory footprint make it a natural candidate for training models with a large number of observations. Due to its popularity, it has been extensively studied in parallel and distributed settings. One standard approach to parallelizing it is the so-called mini-batch SGD framework, where worker nodes compute stochastic gradients on local examples in parallel, and a master node performs the parameter updates. Different variants of this approach have been proposed, both in the synchronous setting \cite{dekel2012optimal}, and the asynchronous setting with delayed updates \cite{agarwal2011distributed}. Another notable work on asynchronous SGD is Hogwild! \cite{hogwild}, where multiple processors carry out  SGD independently and one can overwrite the progress of the other. A caveat of Hogwild! is that it places strong sparsity assumptions on the data. An alternative strategy that is more communication efficient compared to the mini-batch framework is the Parallelized SGD (P-SGD) method \cite{parallelSGD}, which follows the research direction set by \cite{EfficientLargescaleMaxEnt,DistributedPerceptron}. The main idea is to allow each processor to independently perform SGD on the subset of the data that corresponds to it, and then to average all solutions to obtain the final result. Note that in all aforementioned methods, the observations are stored distributively, but not the features.  

\vspace{-2.5mm}
\paragraph*{Coordinate Descent Methods} Coordinate descent methods have proven very useful in various machine learning tasks. In its simplest form, CD selects a single coordinate of the variable vector, and minimizes along that direction while keeping the remaining coordinates fixed \cite{nesterov2012efficiency}. More recent CD versions operate on randomly selected blocks, and update multiple coordinates at the same time \cite{richtarik2014iteration}. Primal CD methods have been studied in the parallel \cite{richtarik2015parallel} and distributed settings \cite{liu2015asynchronous,richtarik2013distributed}. Distributed CD as it appears in \cite{richtarik2013distributed} can be conducted with the coordinates (features) being partitioned, but requires access to all observations. Recently, dual coordinate ascent methods have received ample attention from the research community, as they have been shown to outperform SGD in a number of settings \cite{hsieh2008dual,shalev2013stochastic}. In the dual problem, each dual variable is associated with an observation, so in the distributed setting one would partition the data across observations. Examples of such algorithms include \cite{jaggi2014communication,ma2015adding,yang2013trading}. CoCoA \cite{jaggi2014communication}, which serves as the starting point for D3CA, follows the observation partitioning scheme and treats each block of data as an independent sub-problem. Due to the separability of the problem over the dual variables, the local objectives that are maximized are identical to the global one. Each sub-problem is approximately solved using a dual optimization method; the Stochastic Dual Coordinate Ascent (SDCA) method \cite{shalev2013stochastic} is a popular algorithm for this task. Following the optimization step, the locally updated primal and dual variables are averaged, and the process is repeated until convergence. Similar to SGD-based algorithms, dual methods have not yet been explored when the feature space is distributed.

\vspace{-2.5mm}
\paragraph*{SGD-CD Hybrid Methods}
There has recently been a surge of methods combining SGD
and CD \cite{konevcny2014semi,mokhtari2016doubly,wang2014randomized,xu2015block,zhao2014accelerated}.
These methods conduct parameter updates based on stochastic partial gradients, which are computed by randomly sampling observations and blocks of variables. With the exception of RAPSA \cite{mokhtari2016doubly}, which is a parallel algorithm, all other methods are serial, and typically assume that the sampling process has access to all observations and features. Although this is a valid assumption in a parallel (shared-memory) setting, it does not hold in distributed environments. RAPSA employs an update scheme similar to that of mini-batch SGD, but does not require all variables to be updated at the same time. More specifically, in every iteration each processor randomly picks a subset of observations and a block of variables, and computes a partial stochastic gradient based on them. Subsequently, it performs a single stochastic gradient update on the selected variables, and then re-samples feature blocks and observations. Despite the fact that RAPSA is not a doubly distributed optimization method, its parameter update is quite different from that of RADiSA. On one hand, RAPSA allows only one parameter update per iteration, whereas RADiSA permits multiple updates per iteration, thus leading to a great reduction in communication. Finally, RADiSA utilizes the SVRG technique, which is known to accelerate the rate of convergence of an algorithm.  

\vspace{-2.5mm}
\paragraph*{ADMM-based Methods}
A popular alternative for distributed optimization is the alternating direction method of multipliers (ADMM) \cite{boydadmm}. The original ADMM algorithm, as well as many of its variants that followed (e.g. \cite{mota2013d}), is very flexible in that it can be used to solve a wide variety of problems, and is easily parallelizable (either in terms of features or observations). A block splitting variant of ADMM was recently proposed that allows both features and observations to be stored in distributed fashion \cite{parikh2014block}. One caveat of ADMM-based methods is their slow convergence rate. In our numerical experiments we show empirically the benefits of using RADiSA or D3CA over block splitting ADMM.

\section{Algorithms}

\label{Algorithms}

In this section we present the D3CA and RADiSA algorithms. We first briefly discuss the problem of interest, and then introduce the notation used in the remainder of the paper.

\subsection*{Preliminaries}

\begin{figure*}[h]
\centering
\captionsetup{justification=centering}
\includegraphics[scale=0.7]{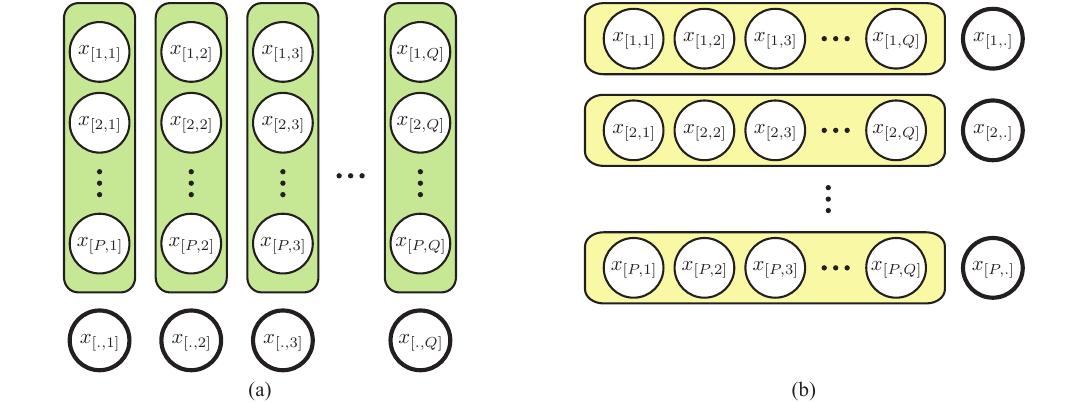}
\caption{An illustration of the partitioning scheme under consideration. (a) and (b) show the definitions of $x_{[.,q]}$ and $x_{[p,.]}$ respectively.}
\label{fig:setupNotation}
\end{figure*}

In a typical supervised learning task, there is a collection of input-output
pairs $\{(x_{i},y_{i})\}_{i=1}^{n}$, where each $x_{i}\in\mathbb{R}^{m}$
represents an observation consisting of $m$ features, and is associated with a corresponding label $y_{i}$. This collection is usually referred to as the training set. The general objective under consideration can be expressed as a minimization problem of a finite sum of convex functions, plus a smooth, convex regularization term (where
$\lambda>0$ is the regularization parameter, and $f_{i}$ is parametrized by $y_{i}$):

\begin{equation}
\label{eq:primal}
\min_{w\in\mathbb{R}^{m}}F(w):={\displaystyle \frac{1}{n}\sum_{i=1}^{n}}f_{i}(w^{T}x_{i})+\lambda||w||^{2}.
\end{equation}

We should remark that additional work would be needed to examine the adaptation of our methods for solving problems with non-smooth regularizers (e.g. $L_{1}$-norm). An alternative approach for finding a solution to \eqref{eq:primal} is to solve its corresponding dual problem. The dual problem of \eqref{eq:primal} has the following form:

\begin{equation}
\label{eq:dual}
\min_{\alpha\in\mathbb{R}^{n}}D(\alpha):={\displaystyle \frac{1}{n}\sum_{i=1}^{n}-\phi^{*}_{i}(-\alpha_{i})-\frac{\lambda}{2}\bigg{|}\hspace{-0.3mm}\bigg{|}\frac{1}{\lambda n}\sum_{i=1}^{n}\alpha_{i}x_{i}\bigg{|}\hspace{-0.3mm}\bigg{|}^{2}},
\end{equation}
where $\phi_{i}^{*}$ is the convex conjugate of $f_{i}$. Note that for certain non-smooth primal objectives used in models such as support vector machines and least absolute deviation, the convex conjugate imposes lower and upper bound constraints on the dual variables. One interesting aspect of the dual objective \eqref{eq:dual} is that there is one dual variable associated with each observation in the training set. Given a dual solution $\alpha\in\mathbb{R}^{n}$, it is possible to retrieve the corresponding primal vector by using 
\begin{equation}
\label{eq:primal-dual}
w(\alpha) = {\displaystyle \frac{1}{\lambda n} \sum_{i=1}^{n}\alpha_{i}x_{i}}. 
\end{equation}

For any primal-dual pair of solutions $w$ and $\alpha$, the duality gap is defined as $F(w)-D(\alpha)$, and it is known that $F(w)\geq D(\alpha)$. Duality theory guarantees that at an optimal solution $\alpha^{*}$ of \eqref{eq:dual}, and $w^{*}$ of \eqref{eq:primal}, $F(w^{*}) = D(\alpha^{*})$.\\


\noindent
\textit{Notation:} We assume that the data $\{(x_{i},y_{i})\}_{i=1}^{n}$ is distributed across observations and features over $K$ computing nodes of a cluster. More specifically, we split the features into $Q$ partitions, and the observations into $P$ partitions (for simplicity we assume that $K=P\cdot Q$). We denote the labels of a partition by $y_{[p]}$, and the observations of the training set for its subset of features by $x_{[p,q]}$. For instance, if we let $Q=2$ and $P=2$, the resulting partitions are $(x_{[1,1]},y_{[1]})$, $(x_{[1,2]},y_{[1]})$, $(x_{[2,1]},y_{[2]})$ and $(x_{[2,2]},y_{[2]})$. Furthermore, $x_{[p,.]}$ represents all observations and features (across all $q$) associated with partition $p$ ($x_{[.,q]}$ is defined similarly) -- Figure~\ref{fig:setupNotation} illustrates this partitioning scheme. We let $n_{p}$ denote the number of observations in each partition, such that $\sum_{p}n_{p}=n$, and we let $m_q$ correspond to the number of features in a partition, such that $\sum_{q}m_{q}=m$. Note that partitions corresponding to the same observations all share the common dual variable $\alpha_{[p,.]}$. In a similar manner, partitions containing the same features share the common primal variable $w_{[.,q]}$. In other words, for some pre-specified values $\tilde{p}$ and $\tilde{q}$, the partial solutions  $\alpha_{[\tilde{p},.]}$ and $w_{[.,\tilde{q}]}$  represent aggregations of the local solutions $\alpha_{[\tilde{p},q]}$ for $q=1,...,Q$ and $w_{[p,\tilde{q}]}$ for $p=1,...,P$. At any iteration of D3CA, the global dual variable vector can be written as $\alpha=[\alpha_{[1,.]},\alpha_{[2,.]},...,\alpha_{[P,.]}]$, whereas for RADiSA the global primal vector has the form $w=[w_{[.,1]},w_{[.,2]},...,w_{[.,Q]}]$, i.e. the global solutions are formed by concatenating the partial solutions.

\subsection*{Doubly Distributed Dual Coordinate Ascent}

The D3CA framework presented in Algorithm~\ref{alg:D3CA} hinges on CoCoA \cite{jaggi2014communication}, but it extends it to cater for the features being distributed as well. The main idea behind D3CA is to approximately solve the local sub-problems using a dual optimization method, and then aggregate the dual variables via averaging. The choice of averaging is reasonable from a dual feasibility standpoint when dealing with non-smooth primal losses -- the \textsc{LocalDualMethod} guarantees that the dual variables are within the lower and upper bounds imposed by the convex conjugate, so their average will also be feasible. Although in CoCoA it is possible to recover the primal variables directly from the local solver, in D3CA, due to the averaging of the dual variables, we need to use the primal-dual relationship to obtain them. Note that in the case where $Q=1$, D3CA reduces to CoCoA. 

\begin{algorithm}[h]
\caption{Doubly Distributed Dual Coordinate Ascent (D3CA)}
\label{alg:D3CA}
\vspace{-2mm}
\begin{flushleft} \textbf{Data: }$(x_{[p,q]},y_{[p]})$  for $p=1,...,P$ and $q=1,...,Q$
\end{flushleft}
\vspace{-5mm}
\begin{flushleft} \textbf{Initialize: $\alpha^{(0)}\leftarrow0$}, \textbf{$w^{(0)}\leftarrow0$}
\end{flushleft}
\vspace{-2mm}
\begin{algorithmic}[1]
\For{$t=1,2,...$}
\For {\textbf{all partitions} $[p,q]$} \textbf{in parallel}
\State $\Delta\alpha_{[p,q]}^{(t)}=$\textsc{LocalDualMethod}$(\alpha^{(t-1)}_{[p,.]},w^{(t-1)}_{[.,q]})$\label{lst:line:localdualstep}
\EndFor
\For {\textbf{all $p$}} \textbf{in parallel} 
\State $\alpha_{[p,.]}^{(t)}=\alpha_{[p,.]}^{(t-1)}+{\frac{1}{P\cdot Q}}{ \sum_{q=1}^{Q}}\Delta\alpha_{[p,q]}^{(t)}$ \label{lst:line:dualave}
\EndFor
\For { \textbf{all} $q$} \textbf{in parallel}
\State $w_{[.,q]}^{(t)}=\frac{1}{\lambda n}{\sum_{p=1}^{P}}((\alpha_{[p,q]}^{(t)})^{T}x_{[p,q]})$\label{lst:line:getPrimal} 
\EndFor
\EndFor
\end{algorithmic}
\end{algorithm}


D3CA requires the input data to be doubly partitioned across $K$ nodes of a cluster. In step~\ref{lst:line:localdualstep}, the algorithm calls the local dual solver, which is shown in Algorithm~\ref{alg:SDCA}. The \textsc{LocalDualMethod} of choice is SDCA \cite{shalev2013stochastic}, with the only difference that the objective that is maximized in step~\ref{lst:line:maxstep} is divided by $Q$. The reason for this is that each partition now contains $\frac{m}{Q}$ variables, so the factor $\frac{1}{Q}$ ensures that the sum of the local objectives adds up to \eqref{eq:dual}. Step~\ref{lst:line:dualave} of Algorithm~\ref{alg:D3CA} shows the dual variable update, which is equivalent to averaging the dual iterates coming from SDCA. Finally, step~\ref{lst:line:getPrimal} retrieves the primal variables in parallel using the primal-dual relationship. The new primal and dual solutions are used to warm-start the next iteration. The performance of the algorithm turns out to be very sensitive to the regularization parameter $\lambda$. For small values of $\lambda$ relative to the problem size, D3CA is not always able to reach the optimal solution. One modification we made to alleviate this issue was to add a step-size parameter when calculating the $\Delta\alpha$'s in the local dual method (Algorithm~\ref{alg:SDCA}, step~\ref{lst:line:maxstep}). In the case of linear Support Vector Machines (SVM) where the closed form solution for step~\ref{lst:line:maxstep} is given by $\Delta\alpha=y_{i}\max(0,\min(1,\frac{\lambda n(1-x_{i}^{T}w^{(h-1)}y_{i})}{||x_{i}||^{2}}+\alpha_{i}^{(h-1)}y_{i}))-\alpha_{i}^{(h-1)}$, we replace $||x_{i}||^{2}$ with a step-size parameter $\beta$ \cite{takavc2013mini}. In our experiments we use $\beta = \frac{\lambda}{t}$, where $t$ is the global iteration counter. Although, a step-size of this form does not resolve the problem entirely, the performance of the method does improve.


\begin{figure*}[h]
\centering
\captionsetup{justification=centering}
\includegraphics[scale=0.9]{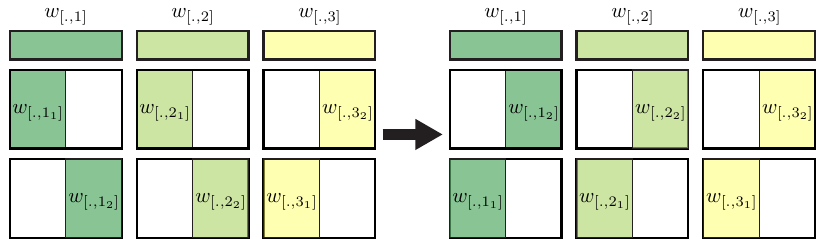}
\caption{An illustration of two iterations of RADiSA, with six overall partitions ($P=2$  and $Q=3$).}
\label{fig:RADiSA}
\end{figure*}

\begin{algorithm}
\caption{\textsc{LocalDualMethod}: Stochastic Dual Coordinate Ascent (SDCA)}
\label{alg:SDCA}
\vspace{-2mm}
\begin{flushleft} \textbf{Input}: $\alpha_{[p,q]}\in\mathbb{R}^{n_{p}}$, $w_{[p,q]}\in\mathbb{R}^{m_{q}}$
\end{flushleft}
\vspace{-5mm}
\begin{flushleft} \textbf{Data: }Local $(x_{[p,q]},y_{[p]})$ 
\end{flushleft}
\vspace{-5mm}
\begin{flushleft} 
\textbf{Initialize: $\alpha^{(0)}\leftarrow\alpha_{[p,q]}$},\textbf{$w^{(0)}\leftarrow w_{[p,q]}$}, $\Delta\alpha_{[p,q]}\leftarrow 0$
\end{flushleft}
\vspace{-3mm}
\begin{algorithmic}[1]
\For {$h=1,2,...$}

\State choose $i\in\{1,2,...,n_{p}\}$ at random

\State find $\Delta\alpha$ maximizing $-{ \frac{1}{Q}}\phi_{i}^{*}(-(\alpha_{i}^{(h-1)}+\Delta\alpha))-$ \label{lst:line:maxstep} 
\item[]\hspace{6mm}${\frac{\lambda n}{2}}||w^{(h-1)}+(\lambda    n)^{-1}\Delta\alpha(x_{[p,q]})_{i}||^{2})$

\State $\alpha_{i}^{(h)}=\alpha_{i}^{(h-1)}+\Delta\alpha$

\State $(\Delta\alpha_{[p,q]})_{i}=(\Delta\alpha_{[p,q]})_{i}+\Delta\alpha$

\State $w^{(h)}=w^{(h-1)}+{ \frac{1}{\lambda n}}\Delta\alpha$$(x_{[p,q]})_{i}$

\EndFor

\State \textbf{Output: $\Delta\alpha_{[p,q]}$}

\end{algorithmic}
\end{algorithm}
In terms of parallelism, the $P\times Q$ sub-problems can be solved independently. These independent processes can either be carried out on separate computing nodes, or in distinct cores in the case of multi-core computing nodes. The only steps that require communication are step~\ref{lst:line:dualave} and step~\ref{lst:line:getPrimal}. The communication steps can be implemented via \textit{reduce} operations -- in Spark we use \textit{treeAggregate}, which is superior to the standard \textit{reduce} operation.

\subsection*{Random Distributed Stochastic Algorithm}

Similar to D3CA, RADiSA, outlined in Algorithm~\ref{alg:RADiSA}, assumes that the data is doubly distributed across $K$ partitions. Before reaching step~\ref{lst:line:outter_for_loop} of the algorithm, all partitions associated with the same block of variables (i.e. $[.,q]$ for $q=1,...,Q$) are further divided into $P$ non-overlapping sub-blocks. The reason for doing this is to ensure that at no time more than one processor is updating the same variables. Although the blocks remain fixed throughout the runtime of the algorithm, the random exchange of sub-blocks  between iterations is allowed (step~\ref{lst:line:sub-block_assignment}). The process of randomly exchanging sub-blocks can be seen graphically in Figure~\ref{fig:RADiSA}. For example, the two left-most partitions that have been assigned the coordinate block $w_{[.,1]}$, exchange sub-blocks $w_{[.,1_{1}]}$ and $w_{[.,1_{2}]}$ from one iteration to the next. The notation $\bar{q}^{q}_{p}$ in step~\ref{lst:line:sub-block_assignment} of the algorithm essentially implies that sub-blocks are partition-specific, and, therefore, depend on $P$ and $Q$. 

A possible variation of Algorithm~\ref{alg:RADiSA} is one that allows for complete overlap between the sub-blocks of variables. In this setting, however, concatenating all local variables into a single global solution (step~\ref{lst:line:sol_concat}) is no longer an option. Other techniques, such as parameter averaging, need to be employed in order to aggregate the local solutions. In our numerical experiments, we explore a parameter averaging version of RADiSA (RADiSA-avg).

The optimization procedure of RADiSA makes use of the Stochastic Variance Reduce Gradient (SVRG) method \cite{johnson2013accelerating}, which helps accelerate the convergence of the algorithm. SVRG requires a full-gradient computation (step~\ref{lst:line:full_gradient}), typically after a full pass over the data. Note that for models that can be expressed as the sum functions, like in \eqref{eq:primal}, it is possible to compute the gradient when the data is doubly distributed. Although RADiSA by default computes a full-gradient for each global iteration, delaying the gradient updates can be a viable alternative. Step~\ref{lst:line:svrg_update} shows the standard SVRG step,\footnote{In StepStep~\ref{lst:line:svrg_update}, $x_{[p,\bar{q}]_{j}}$ corresponds to the features in the $j^{th}$ row of sub-block $\bar{q}$ in partition $[p,q]$.} which is applied to the sub-block of coordinates assigned to that partition. The total number of inner iterations is determined by the batch size $L$, which is a hyper-parameter. As is always the case with variants of the SGD algorithm, the learning rate $\eta_{t}$ (also known as step-size) typically requires some tuning from the user in order to achieve the best possible results. In Section ~\ref{NumExperiments} we discuss our choice of step-size. The final stage of the algorithm simply concatenates all the local solutions to obtain the next global iterate. The new global iterate is used to warm-start the subsequent iteration. 

Similar to D3CA, the $P\times Q$ sub-problems can be solved independently. As far as communication is concerned, only the gradient computation (step~\ref{lst:line:full_gradient}) and parameter update (step~\ref{lst:line:svrg_update}) stages require coordination among the different processes. In Spark, the communication operations are implemented via $treeAggregate$.

\begin{algorithm}[h]
\caption{Random Distributed Stochastic Algorithm (RADiSA)}
\label{alg:RADiSA}
\vspace{-2mm}
\begin{flushleft} \textbf{Input: } batch size $L$, learning rate $\eta_{t}$
\end{flushleft}
\vspace{-5mm}
\begin{flushleft} \textbf{Data: }$(x_{[p,q]},y_{[p]})$  for $p=1,...,P$ and $q=1,...,Q$ 
\end{flushleft}
\vspace{-5mm}
\begin{flushleft} 
\textbf{Initialize: $\tilde{w}_{0}\leftarrow0$} 
\end{flushleft}
\vspace{-5mm}
\begin{flushleft} 
Partition each $[.,q]$ into $P$ blocks, such that  $w_{[.,q]}=[w_{[.,q_{1}]},w_{[.,q_{2}]},...,w_{[.,q_{P}]}]$
\end{flushleft}
\vspace{-3mm}
\begin{algorithmic}[1]
\For{$t=1,2,...$} \label{lst:line:outter_for_loop}
\State $\tilde{w}=\tilde{w}^{(t-1)}$ 
\State $\tilde{\mu}={ \frac{1}{n}\sum_{i=1}^{n}\nabla f_{i}(}\tilde{w}^{T}x_{i})$ \label{lst:line:full_gradient}
\For {\textbf{all partitions } $[p,q]$} \textbf{in parallel} \label{lst:line:opt_for}
\vspace{1mm}
\State Randomly pick sub-block $\bar{q} = \bar{q}^{q}_{p}$ in non-overlapping manner\label{lst:line:sub-block_assignment} 
\State $w^{(0)}=\tilde{w}_{[p,\bar{q}]}$
\For {$i=0,...,L-1$}
\State randomly pick $j\in\{1,...,n_{p}\}$
\State $w^{(i+1)}=w^{(i)}-\eta_{t}(\hat{\nabla}f_{j}(w^{(i)T}x_{[p,\bar{q}]_{j}})-\hat{\nabla}f_{j}(\tilde{w}_{[p,\bar{q}]}^{T}x_{[p,\bar{q}]_{j}})+\tilde{\mu}_{[p,\bar{q}]})$\label{lst:line:svrg_update}
\EndFor 
\EndFor \label{lst:line:end_opt_for}
\State $\tilde{w}^{(t)} = [w_{[.,1]},w_{[.,2]},...,w_{[.,Q]}]$ \label{lst:line:sol_concat}, where ${w}_{[.,q]}=[{w}^{(L)}_{[.,\bar{q}^{q}_{1}]},...,{w}^{(L)}_{[.,\bar{q}^{q}_{P}]}]$
\EndFor
\end{algorithmic}
\end{algorithm}





\section{Numerical Experiments}
\label{NumExperiments}

\begin{figure*}[h]
\centering
\includegraphics[scale=0.65]{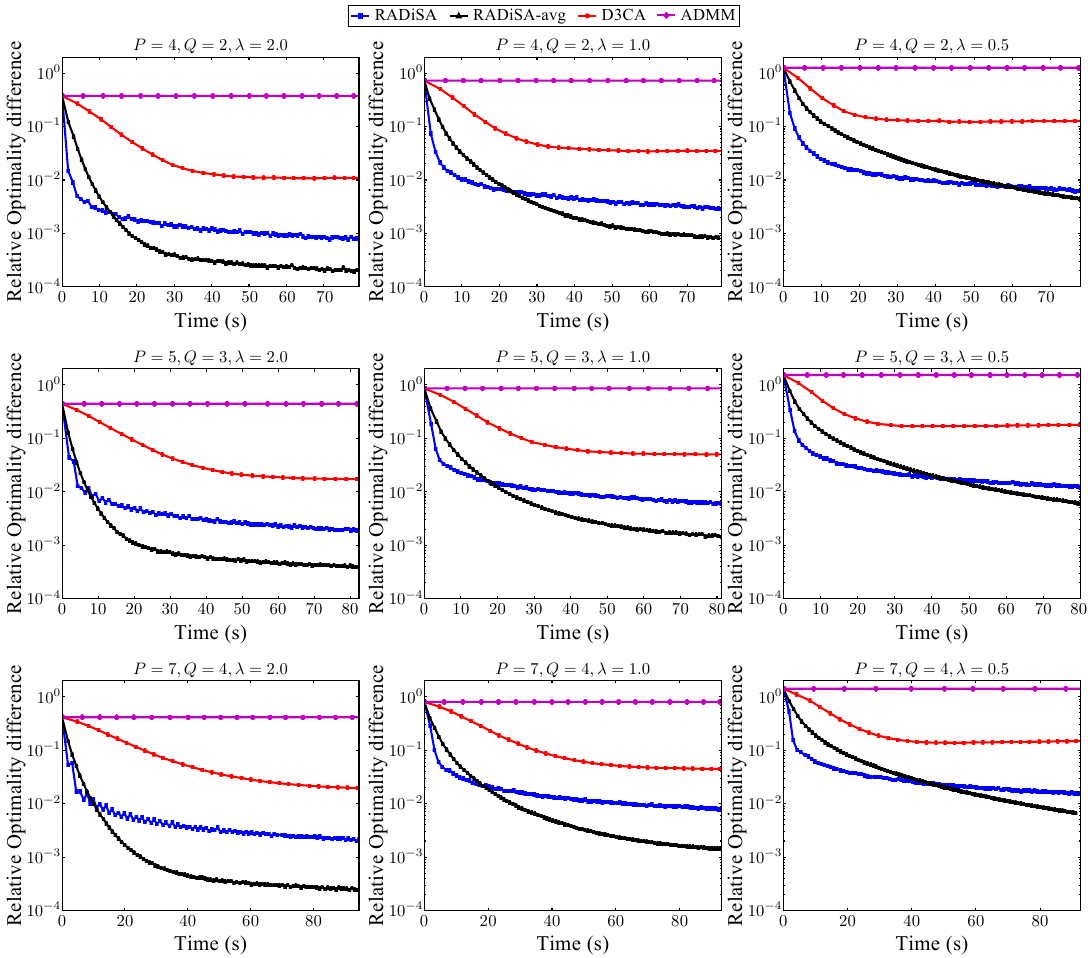}
\caption{Relative optimality difference against elapsed time for three data sets with the following configurations of $P$ and $Q$: (4,2), (5,3) and (7,4).} 
\label{fig:experiments1}
\end{figure*}

\vspace{-3mm}

\begin{figure*}[h]
\centering
\includegraphics[scale=0.65]{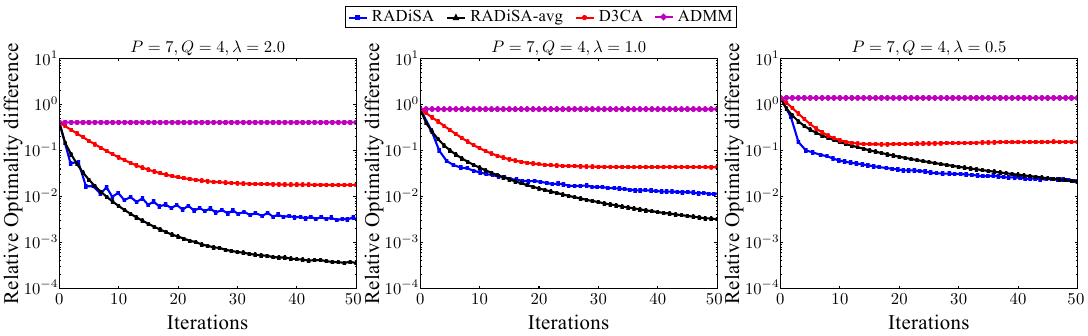}
\caption{Relative optimality difference against iteration count.} 
\label{fig:experiments1_5}
\end{figure*}


In this section we present two sets of experiments. The first set is adopted from \cite{parikh2014block}, and we compare the block distributed version of ADMM with RADiSA and D3CA. In the second set of experiments we explore the scalability properties of the proposed methods. We implemented all algorithms in Spark and conducted the experiments in a Hadoop cluster with 4 nodes, each containing 8 Intel Xeon E5-2407 2.2GHz cores. For the ADMM method, we follow the approach outlined in \cite{parikh2014block}, whereby the Cholesky factorization of the data matrix is computed once, and is cached for re-use in subsequent iterations. Since the computational time of the Cholesky decomposition depends substantially on the underlying BLAS library, in all figures reporting the execution time of ADMM, we have excluded the factorization time. This makes the reported times for ADMM lower than in reality. 

The problem solved in \cite{parikh2014block} was lasso regression, which is not a model of the form \eqref{eq:primal}. Instead, we trained one of the most popular classification models: binary classification hinge loss support vector machines (SVM). The data for the first set of experiments was generated according to a standard procedure outlined in \cite{umichadmm}: the $x_{i}$'s and $w$ were sampled from the $[-1,1]$ uniform distribution; $y_{i}=\text{sgn}({w^{T}x_{i}})$, and the sign of each $y_{i}$ was randomly flipped with probability 0.1. The features were standardized to have unit variance. We take the size of each partition to be dense $2,000\times3,000$,\footnote{In \cite{parikh2014block} the size of the partitions was $3,000\times 5,000$, but due to the BLAS issue mentioned earlier, we resorted to smaller problems to obtain comparable run-times across all methods.} and set $P$ and $Q$ accordingly to produce problems at different scales. For example, for $P=4$ and $Q=2$, the size of the entire instance is $8,000\times6,000$. The information about the three data sets is summarized in table~\ref{tab:Data1}. As far as hyper-parameter tuning is concerned, for ADMM we set $\rho=\lambda$. For RADiSA we set the step-size to have the form $\eta_{t}=\frac{\gamma}{(1+\sqrt{t-1})}$, and select the constant $\gamma$ that gives the best performance.

To measure the training performance of the methods under consideration, we use the relative optimality difference metric, defined as $(f^{(t)}-f^{*})/f^{*},$ where $f^{(t)}$ is the primal objective function value at iteration $t$, and $f^{*}$ corresponds to the optimal objective function value obtained by running an algorithm for a very long time.

\vspace{-2mm}

\begin{table}[h]
\caption{Datasets for Numerical Experiments (Part 1)}
\label{tab:Data1} 
\noindent \centering{}%
\begin{tabular}{c|r|r|r}
\textbf{$P\times Q$} & $4\times2$ & $5\times3$ & $7\times4$ \tabularnewline
\hline 
Nonzero entries & $48$M & $90$M & $168$M \tabularnewline
Number of cores used  & $8$ & $15$ & $28$ \tabularnewline
\end{tabular}
\end{table}

\vspace{-2mm}

In Figure~\ref{fig:experiments1}, we observe that RADiSA-avg performs best in all cases, with RADiSA coming in a close second, especially for smaller regularization values. Both variants of RADiSA and D3CA clearly outperform ADMM, which needs a much larger number of iterations to produce a satisfactory solution. We provide an additional comparison in Figure~\ref{fig:experiments1_5} that further demonstrates this point. We plot the relative optimality difference across 50 iterations. One note about RADiSA-avg is that its performance depends heavily on the number of observation partitions. The averaging step tends to dilute the updates, leading to a slower convergence rate. This is evident when training models on larger data sets than the ones shown in this round of experiments. Another important remark we should make is that when dealing with larger data sets, the behavior of D3CA is erratic for small regularization values. For large regularization values, however, it can produce good solutions. 

In the second set of experiments we study the strong scaling properties of our algorithms. Note that the goal of these experiments is to gain insight into the properties of the two methods, rather than to find the best partitioning strategy. The reason for this is that the partitioning of the data is dictated by the application, and is, therefore, out of the practitioner's control. The model under consideration is again linear SVM. To conduct strong scaling experiments, the overall size of the data set does not change, but we increase the number of available computing resources. This means that as the overall number of partitions $K$ increases, the workload of each processor decreases. For RADiSA, we keep the overall number of data points processed constant as we increase $K$, which implies that as the sub-problem/partition size decreases, so does the batch size $L$. One matter that requires attention is the step-size parameter. For all SGD-based methods, the magnitude of the step-size $\eta_{t}$ is inversely proportional to the batch size $L$. We adjust the step-size as $K$ increases by simply taking into account the number of observation partitions $P$. D3CA does not require any parameter tuning. We test our algorithms on two real-world data sets that are available through the LIBSVM website.\footnote{\url{http://www.csie.ntu.edu.tw/~cjlin/libsvmtools/datasets/binary.html}} Table~\ref{tab:Data2} summarizes the details on these data sets. 

\vspace{-2mm}

\begin{table}[h]
\caption{Datasets for Numerical Experiments (Part 2 - Strong Scaling)}
\label{tab:Data2} 
\noindent \centering{}%
\begin{tabular}{c|r|r|r}
Dataset & Observations & Features & Sparsity \tabularnewline
\hline 
real-sim & 72,309 & 20,958 & 0.240\% \tabularnewline
news20 & 19,996 & 1,355,191 & 0.030\% \tabularnewline
\end{tabular}
\end{table}

\vspace{-2mm}

\begin{figure*}[h]
\centering
\includegraphics[scale=0.65]{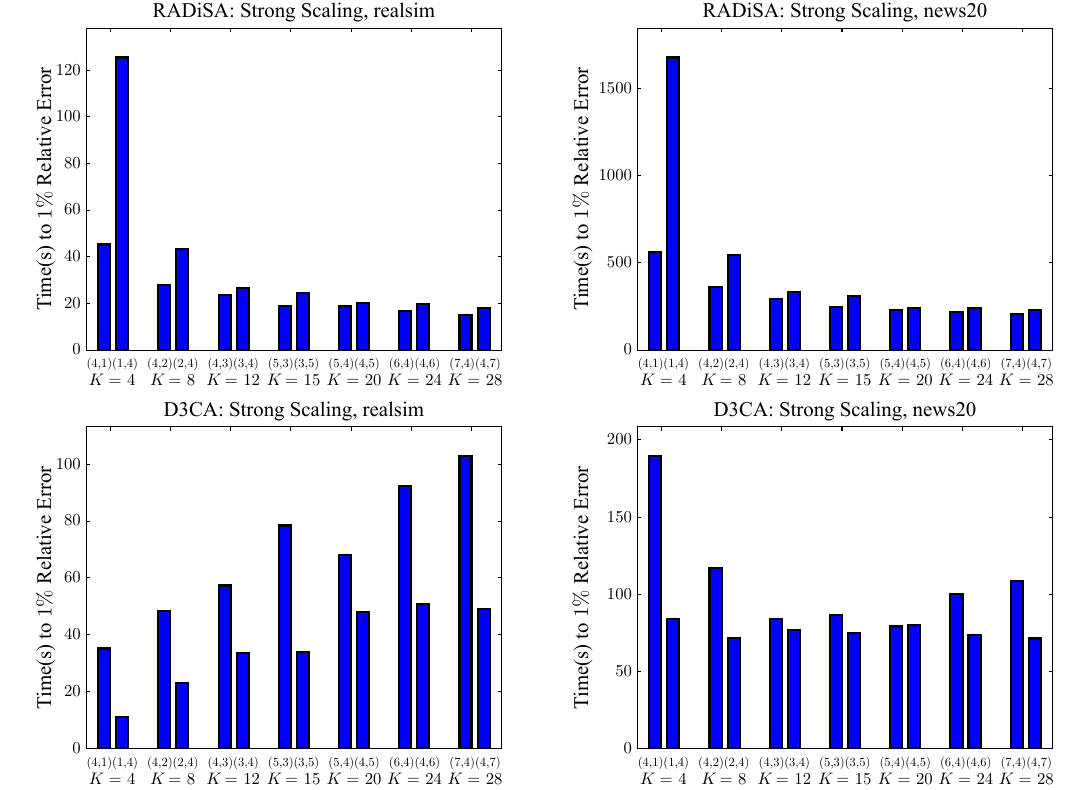}
\caption{Strong scaling of realsim and news20. The $x$-axis shows the various partition configurations for each level of $K$. The $y$-axis shows the total time in seconds that is needed to reach a 1\% optimality difference. The run-time for the two methods is not comparable due to different regularization values being used. For RADiSA we used $\lambda=10^{-3}$ and for D3CA we used $\lambda=10^{-2}$.}
\vspace{-5mm}
\label{fig:experiments2}
\end{figure*}

As we can see in Figure~\ref{fig:experiments2}, RADiSA exhibits strong scaling properties in a consistent manner. In both data sets the run-time decreases significantly when introducing additional computing resources. It is interesting that early configurations with $P<Q$ perform significantly worse compared to the alternate configurations where $P>Q$. Let us consider the configurations (4,1) and (1,4). In each case, the number of variable sub-blocks is equal to $4$. This implies that the dimensionality of the sub-problems is identical for both partition arrangements. However, the second partition configuration has to process four times more observations compared to the first one, resulting in an increased run-time. It is noteworthy that the difference in performance tails away as the number of partitions becomes large enough. Overall, to achieve consistently good results, it is preferable that $P>Q$.

The strong scaling performance of D3CA is mixed. For the smaller data set (realsim), introducing additional computing resources deteriorates the run-time performance. On the larger data set (news20), increasing the number of partitions $K$ pays dividends when $P>Q$. On the other hand, when $Q>P$, providing additional resources has little to no effect. The pattern observed in Figure~\ref{fig:experiments2} is representative of the behavior of D3CA on small versus large data sets (we conducted additional experiments to further attest this). It is safe to conclude that when using D3CA, it is desirable that $Q>P$.

\section{Conclusion}

In this work we presented two doubly distributed algorithms for large-scale machine learning. Such methods can be particularly flexible, as they do not require each node of a cluster to have access to neither all features nor all observations of the training set. It is noteworthy that when massive datasets are already stored in a doubly distributed manner, methods such as the ones introduced in this paper may be the only viable option. Our numerical experiments show that both methods outperform the block distributed version of ADMM. There is, nevertheless, room to improve both methods. The most important task would be to derive a step-size parameter for D3CA that will guarantee the convergence of the algorithm for all regularization parameters. Furthermore, removing the bottleneck of the primal vector computation would result into a significant speedup. As far as RADiSA is concerned, one potential extension would be to incorporate a streaming version of SVRG \cite{frostig2014competing}, or a variant that does not require computation of the full gradient at early stages \cite{babanezhad2015stop}. Finally, studying the theoretical properties of both methods is certainly a topic of interest for future research. 

%
%
{\small\bibliographystyle{abbrv}
\bibliography{master}}

\end{document}